\documentclass[10pt, conference]{IEEEtran}

\usepackage{graphicx}

\usepackage{amssymb}

\begin{document}





\title{Automated Image Processing for the Analysis of\\ DNA Repair Dynamics}

\author{\IEEEauthorblockN{Thorsten~Rie\ss\IEEEauthorrefmark{1},
Christian~Dietz\IEEEauthorrefmark{1}\IEEEauthorrefmark{2},
Martin~Tomas\IEEEauthorrefmark{2}\IEEEauthorrefmark{3}, 
Elisa~Ferrando-May\IEEEauthorrefmark{3} and
Dorit~Merhof\IEEEauthorrefmark{1}\IEEEauthorrefmark{4}}
\IEEEauthorblockA{\IEEEauthorrefmark{1}Interdisciplinary Center for Interactive Data Analysis, Modelling and Visual Exploration (INCIDE)}
\IEEEauthorblockA{\IEEEauthorrefmark{2}Department of Physics, Center for Applied Photonics (CAP)}
\IEEEauthorblockA{\IEEEauthorrefmark{3}Department of Biology, Bioimaging Center (BIC)}
\IEEEauthorblockA{\IEEEauthorrefmark{4}Visual Computing\\ University of Konstanz, 78457 Konstanz, Germany\\ E-Mail: \{thorsten.riess, christian.dietz,
martin.tomas, elisa.may, dorit.merhof\}@uni-konstanz.de}
}

\maketitle
\begin{abstract}

The efficient repair of cellular DNA is essential for the maintenance and
inheritance of genomic information.  In order to cope with the high frequency of
spontaneous and induced DNA damage, a multitude of repair mechanisms have
evolved.  These are enabled by a wide range of protein factors specifically
recognizing different types of lesions and finally restoring the normal DNA
sequence.  This work focuses on the repair factor XPC (xeroderma pigmentosum
complementation group C), which identifies bulky DNA lesions and initiates their
removal via the nucleotide excision repair pathway.  The binding of XPC to
damaged DNA can be visualized in living cells by following the accumulation of a
fluorescent XPC fusion at lesions induced by laser microirradiation in a
fluorescence microscope.

In this work, an automated image processing pipeline is presented which allows
to identify and quantify the accumulation reaction without any user interaction.
The image processing pipeline comprises a preprocessing stage where the image
stack data is filtered and the nucleus of interest is segmented.  Afterwards, the
images are registered to each other in order to account for movements of the
cell, and then a bounding box enclosing the XPC-specific signal is automatically
determined.  Finally, the time-dependent relocation of XPC is evaluated by
analyzing the intensity change within this box.

Comparison of the automated processing results with the manual evaluation yields
qualitatively similar results. However, the automated analysis provides more
accurate, reproducible data with smaller standard errors.

The image processing pipeline presented in this work allows for an efficient
analysis of large amounts of experimental data with no user interaction
required.

\end{abstract}


\IEEEpeerreviewmaketitle



\section{Introduction}
\label{sec:intro}

Damage to the DNA of cells occurs either due to environmental factors
\emph{(exogenous damage)} or due to natural metabolic processes
\emph{(endogenous damage)}. Exogenous damage may be caused by exposure to UV
light or other types of radiation including $\gamma$-rays, or toxins and
chemicals.  Endogenous damage is mostly caused by reactive oxygen species
produced from normal metabolic byproducts, and also includes replication errors
during mitosis.

DNA damage occurs at a rate of 1,000 to 1,000,000 molecular lesions per cell per
day~\cite{Lodish04}, affecting only 0.000165\% of the human genome's
approximately 6 billion bases (3 billion base pairs). However, unrepaired damage
in critical genes (such as tumor suppressor genes) can interfere with normal
cell physiology and increase the likelihood of tumor formation significantly.

Nucleotide excision repair (NER)~\cite{Gillet06} is a fundamental protective
system that promotes genome stability by eliminating a wide range of DNA
lesions.  Transcription-coupled repair (TCR), which takes place when the
transcription machinery is blocked by obstructing lesions in the transcribed
strand~\cite{Hanawalt08}, and global genome repair (GGR) are the two alternative
mechanisms of the NER pathway.

The xeroderma pigmentosum group C (XPC) protein is an important protein involved
in the GGR pathway. It recognizes DNA damage and initiates the DNA excision
repair of helix-distorting base lesions. In healthy cells, the damage is excised
by endonucleases, the missing sequence is replaced by DNA polymerase, and a ligase
completes the reaction.

An important research question in the field of DNA repair concerns the
mechanisms by which the sensor-like protein XPC actually finds base lesions
among a large excess of native DNA in a typical mammalian
genome~\cite{Schaerer07,Sugasawa07}. One approach to investigate how XPC
searches for aberrant sites within the DNA consists in the visualization of the
time-dependent relocation of fluorescently labeled XPC to sites of DNA damage
induced at high spatial resolution by irradiation with a femtosecond
laser~\cite{Camenisch09}.  For this purpose, XPC was marked with green
fluorescent protein (GFP), which allows to investigate the damage-dependent
recruitment of the fusion product XPC-GFP by confocal microscopy.

In the analysis pursued in~\cite{Camenisch09} the accumulation of XPC-GFP at
the induced lesions was quantified by manually defining a bounding box enclosing the
lesion and measurement of the intensity change due to accumulation of
XPC-GFP in this box over time.

From an image processing point of view, such a manual analysis is unsatisfactory
and error-prone for several reasons: First of all, a manual analysis of a
significant amount of mammalian cells is a tedious and time consuming task for
the investigator. Furthermore, the results obtained from a manual analysis are
not rater-independent and lack robustness and reproducibility.

The challenges for implementing an automated image processing pipeline are due
to the low resolution of the image stacks, due to movements of the cells over
time, due to other obscuring cells and structures, and due to the low contrast
between the DNA damage and the surrounding nucleus.  An overview of current
methods for the analysis of fluorescent microscopy images can be found
in~\cite{Gitai09} and references therein.

In this work, these issues are addressed and an automated image processing
approach is proposed.  This approach allows for automatically detecting the
region of XPC accumulation on image stacks showing the cell nuclei and for evaluating the
damage-induced changes of XPC dynamics in the nuclear compartment over time.
The approach comprises a preprocessing stage, where the image stack data is
filtered and the nucleus under consideration is segmented in each image.  The
images are then registered to each other in order to account for movements of
the cell and a bounding box enclosing the DNA damage is automatically
determined.  Finally, the time-dependent relocation of XPC is evaluated by
analyzing the intensity change due to accumulation of XPC-GFP.

\section{Material and Methods}
\label{sec:methods}

In this section, the image acquisition and the image processing methods for
evaluating the time-dependent relocation of XPC-GFP are presented.  The
microscopy image stacks are acquired in an experimental setup explained in
Section~\ref{sec:data}.  The image stacks are then imported into the software
framework \textsc{KNIME}, which is described in section~\ref{sec:software},
where the image processing pipeline is implemented.  This pipeline consists of a
segmentation and registration step as shown in section~\ref{sec:segmentation}
and the detection of the region of interest (ROI).  A scoring algorithm for
this detection is presented in section~\ref{sec:detection}.  Finally, the pixel
intensity measurement within the ROI is explained in
section~\ref{sec:measurement}.

\subsection{Biological model system and image data acquisition}
\label{sec:data}

Recruitment of the DNA repair factor XPC to sites of DNA damage was monitored in
live cells by confocal microscopy.  To this end fluorescent fusions of wildtype
(WT) XPC and of various XPC mutants (F733A, F756A, F797A, F799A, N754A, W531A,
W542A, W690A, W690S) were expressed in either Simian virus $40$-transformed human
XP-C fibroblasts or in Chinese Hamster Ovary cells (CHO).  Cells were then
irradiated at the microscope stage (Zeiss LSM 5 Pascal) using an in-house built
femtosecond Er:fiber laser focused through a 40x oil immersion objective lens.
DNA lesions were induced at $775nm$ by multiphoton
absorption~\cite{Traeutlein08}.  A cell nucleus expressing XPC-GFP was placed in
the center of the field of view and imaged prior to and at defined time
intervals after femtosecond laser irradiation. XPC-GFP fluorescence was detected
using a $488nm$ Ar-laser.  The acquired image stacks consist of one
pre-irradiation frame, one dark frame recorded while scanning with the fiber
laser and $60$ or $52$ post-irradiation frames for experiments with XP-C cells and
CHO cells, respectively.  The frames were acquired at time steps of $6$--$7$
seconds and the frame size is either 512x512 pixels or 580x580 pixels. The
femtosecond laser was scanned along a vertical line of $10\mu m$ in length
\cite{Camenisch09,Clement10}.

\subsection{Software framework}
\label{sec:software}

The software platform \textsc{KNIME} (The Konstanz Information
Miner~\cite{Berthold08}) is an open-source tool for data integration,
processing, analysis and exploration. Essentially, \textsc{KNIME} is designed
to import, transform and visualize large data sets in a convenient and easy to
use way. \textsc{KNIME} workflows consist of interacting nodes, which may each
represent an algorithm, a single import routine or a visualization tool.  The
exchange of data between nodes is accomplished via data tables which are passed
from one node to another by node connections. The graphical user interface makes
it possible to construct workflows consisting of different nodes and their
interconnection via a simple drag-and-drop mechanism.  The data flow is visually
represented by connections between the nodes, typically starting with a node to
import the data, followed by one or more processing nodes and finally one or
more output nodes.  Recently, \textsc{KNIME} has been extended to provide basic
image processing nodes such as image input/output and standard thresholding
algorithms.

In this work, \textsc{KNIME} is used as a basis to implement a fully automated
system that measures fluorescence and quantifies the acccumulation of XPC-GFP.
The image processing workflow consists of several custom \textsc{KNIME} nodes
that are combined with standard image processing nodes. This concept allows to
batch process large amounts of image stacks and automatically save the results.
Additionally, due to the modular design of \textsc{KNIME} workflows, it is
possible to assess intermediate results at every stage of the processing
pipeline.

\subsection{Segmentation and registration}
\label{sec:segmentation}

The image processing steps required to quantify the accumulation
kinetics of XPC comprise the segmentation of the nucleus, a registration step,
and the identification of the irradiated area (the ROI). 

In order to identify the nucleus of interest (NOI) in each frame of the image
stack, the image is smoothed using a standard Median filter with radius three,
and an Otsu thresholding algorithm~\cite{Otsu79} is applied.  This results in a
coarse segmentation of each NOI candidate. Since the region scanned during
image acquisition is adjusted such that the NOI is located at the center of the
image, the NOI can be identified even if multiple cell nuclei are visible in
the image by comparing the center of gravity of the NOI candidates.  The center
of gravity of the NOI is then used to create a polar image~\cite{Bamford98,
Kvarnstrom08}, which is convolved with a Gaussian Blur filter with $\sigma=2$
and a standard Median filter of radius three. Then Otsu thresholding is applied
to the filtered polar image, which results in a binary image separating the
nucleus from the background.  In this binary image, the contour of the nucleus
can easily be detected.  Finally, the original image is masked using this
contour, which concludes the segmentation process.  The usage of the polar
image improves the segmentation result over pure smoothing and thresholding the
cartesian image.  To get the same result on the cartesian image only, one would
need a much more elaborate pipeline than in the proposed approach.

\begin{figure}
  \begin{center}
    \includegraphics[width=\columnwidth]{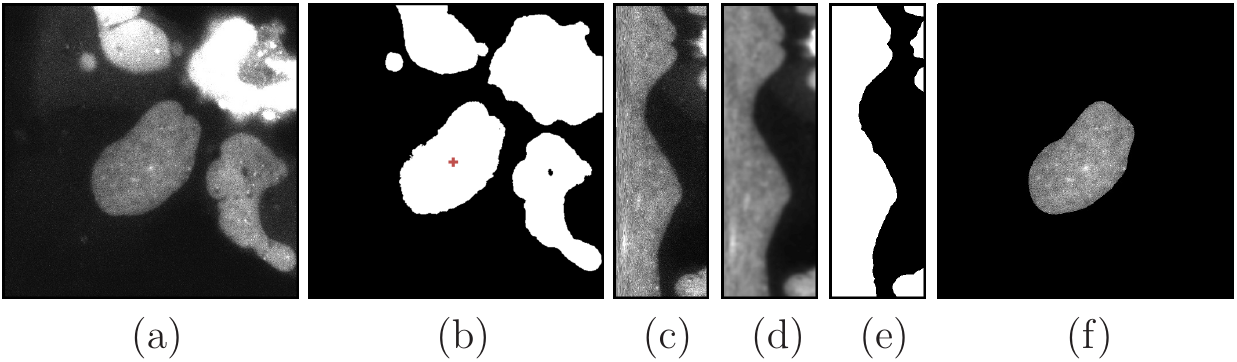}
  \end{center} 
  \caption{Segmentation process for an exemplary image. {\it (a)} Original image
  with the nucleus of interest (NOI) in the middle and surrounding (undamaged)
  objects.  {\it (b)} Binary image after a Otsu thresholding, the center of gravity
  of the NOI is marked. {\it (c)} Polar image. {\it (d)} Smoothed polar image.
  {\it (e)}  Binary image resulting from thresholding the smoothed polar image.
  {\it (f)} Masked original image (segmentation result).
  \label{fig:segmentation}}
\end{figure}

Figure~\ref{fig:segmentation} illustrates the different steps of the
segmentation process and shows the original image including the NOI and
surrounding nuclei {\it (a)}, then the Otsu-thresholded image with the marked
center of gravity {\it (b)} and the polar image {\it (c)}.  Panel {\it (d)}
shows the smoothed polar image, which is again thresholded and the binary image
is shown in panel {\it (e)}.  The final result is the masked original image
shown in panel {\it (f)}.  The resulting segmentation is very accurate and
suitable for further processing steps.

In the next step, registration of the individual frames is required due to
potential movement and deformation of the nucleus over time. For this purpose,
a rigid body registration algorithm~\cite{Thevenaz1998} is applied to the
cropped images from the previous segmentation step.  As a result, the position
of the NOI remains the same across all frames, and also the area of induced
damage remains stationary.  This allows us to determine a single ROI which
comprises the area of intensity change due to accumulation of XPC-GFP for all
image frames in the stack.  In very rare cases the nucleus deforms during image
acquisition and this deformation also affects the treated area.  The rigid body
registration does not compensate for this deformation, hence the ROI that is
used in the following step is chosen slightly larger than necessary if the
nucleus does not deform, but still gives a sufficient accuracy.

\subsection{Detection of the area of XPC-GFP accumulation}
\label{sec:detection}

In order to detect the ROI, a time-averaging projection, avg-$t$-projection for
short, of the registered image stack is created in a first step.  In this
projection, the damaged area is expected to show up as a vertical line of high
intensity.  In some example image stacks, this area can be clearly identified
also by a non-expert, in others it is at the limit of visual detection.  The
ROI is a box of fixed size that exactly covers the irradiated area; for the
automatic detection the user can adjust the exact width and height that is used
in the algorithm.  The box orientation is always such that the long side is
vertical, since the irradiated area is a vertical line segment, and the
movement/deformation of the nucleus after registration is negligible.

For the automated detection of the correct region, a sum-$y$-projection is
applied to the avg-$t$-projection image, which sums up the intensity values for
each column of the image.  This results in an intensity profile of the columns,
where the damaged area is expected to show up as a peak.  A Median filter with
radius three is applied to this intensity profile for smoothing.  Due to noise,
poor image quality and additional bright spots in the nucleus, this peak is not
unique and not straight-forward to identify in some experimental images. For
this reason, a combination of three different scoring methods is employed to
detect the correct peak.  The first scoring method is the height of the peak
compared to the neighbouring peaks.  The second score is obtained by
calculating the response of a ``Haar-like'' feature~\cite{Lienhart03} centered
at the peak and with a fixed width slightly larger than the width of the ROI.
This can be interpreted as locally measuring the difference of the area
underneath the peak with neighbouring areas.  Finally, the third score is the
distance of the peak to the center of the nucleus.  In an ideal experiment, the
irradiation is applied exactly at the center of the NOI, and hence a peak close
to the center is more likely to be the correct one.  For the automatic
evaluation all three scores are weighted equally and the peak with the highest
overall score is chosen.

After the correct peak has been identified, the ROI is chosen such that it is
centered at the $x$-position of the peak and has the predefined width and
height.  If the height of the nuclues exceeds the height of the ROI, the
$y$-position of the ROI is adjusted such that the intensity in the
avg-$t$-projection is maximized.  Examples are shown in
Figure~\ref{fig:roidetection} below.

\subsection{Intensity measurement}
\label{sec:measurement}

The time-dependent relocation of XPC is evaluated by analyzing the intensity
change due to accumulation of XPC-GFP.  The measurement is performed on the
pre-processed and registered image stack and is accomplished as follows: First,
the background is subtracted from the ROI for each image frame in order to
improve the results of the intensity measurement.  The average pixel intensity
of the region of interest $I_{ROI}$, as well as the average pixel intensity
value $I_{NOI}$ of the whole NOI, is then measured at every time-step.  The
quotient $I_{ROI}/I_{NOI}$ of these values is computed and scaled such that the
value of the first image (pre-irradiation) is always equal to one.

In order to validate the presented approach, time-lapse series of a total of
$11$ different XPC mutants have been analyzed, with each series repeated at
least $8$ times.  The XPC mutants were expressed in two different cell types,
human XPC fibroblasts and CHO cells. 

The measurement values for each mutant and each cell type were averaged at each
time-step and the standard errors were computed, which allows easy comparison to
the manual evaluation described in~\cite{Camenisch09,Clement10}.

\section{Results}

The results of the automated measurement are presented in the following.  In
particular, the segmentation performance is evaluated in
section~\ref{sec:performance}, and the measurement results are compared to the
manual evaluation in section~\ref{sec:comparison}.

\subsection{Segmentation Performance}
\label{sec:performance}

\begin{figure}
  \begin{center}
    \includegraphics[width=\columnwidth]{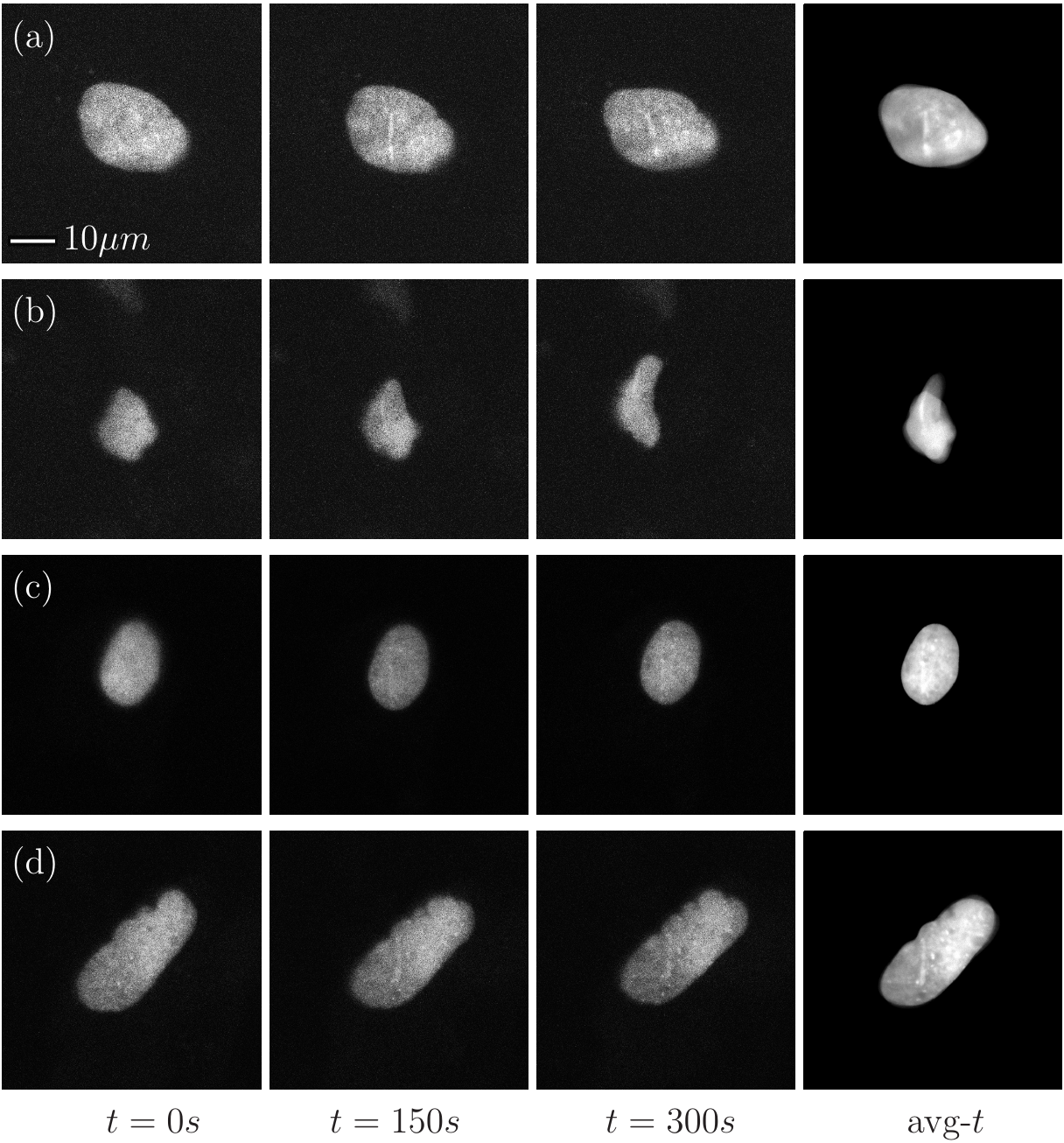}
  \end{center}
  \caption{Microscopy images.  Each row shows four images, the first one
  recorded at the beginning of the experiment (pre-irradiation), the second
  after $150$ seconds and the third after $300$ seconds.  The fourth image shows
  the avg-$t$ projection \emph{after} the segmentation and registration
  algorithm has been applied.  The XPC-GFP accumulation within the nucleus in
  row {\it (a)} is clearly visible.  The nucleus in row {\it (b)} deforms and
  moves during image acquisition.  Row {\it (c)} shows a nucleus where the
  accumulation is hardly detectable.  The whole nucleus in row {\it (d)} is very
  bright, hence the XPC-GFP accumulation is hardly visible.  Note that all
  microscopy images shown here are contrast enhanced.\label{fig:examples}}
\end{figure}

\begin{figure}
  \begin{center}
    \includegraphics[width=\columnwidth]{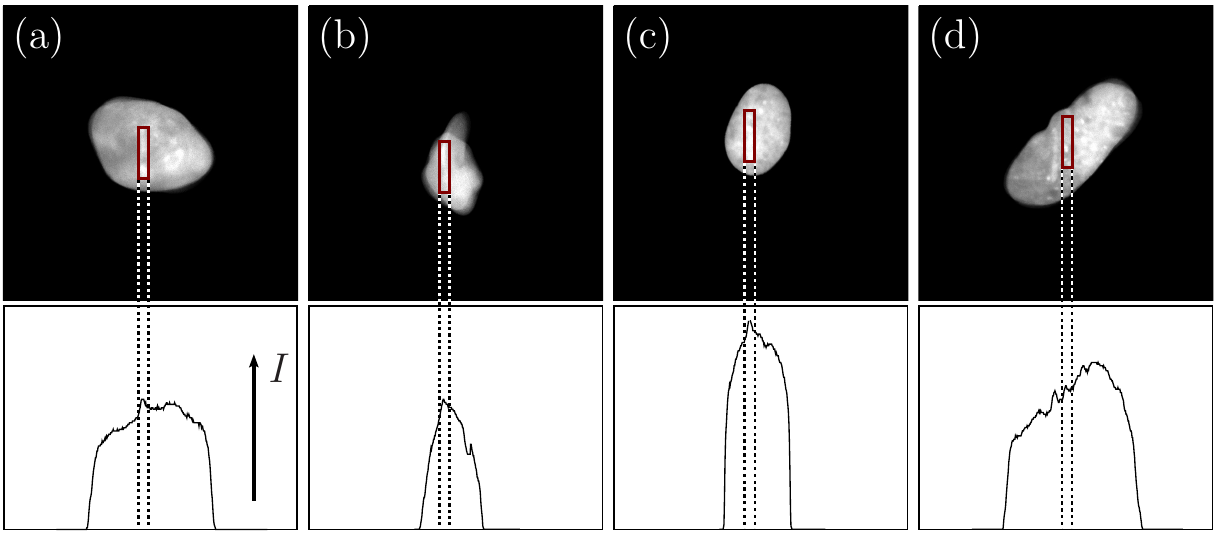}
  \end{center}
  \caption{The detection of the ROI.  The top row shows representative images of
  avg-$t$-projections, the bottom row shows the associated intensity profile,
  the intensity is denoted by $I$.  The detected ROI is marked in each
  avg-$t$-projection image, as well as the associated peak in the intensity
  profile.  The panels {\it (a)}--{\it (d)} correspond to the images shown in
  Figure~\ref{fig:examples} {\it (a)}--{\it (d)}.  Note that panel {\it (d)}
  shows a negative example: the irradiated area lies to the left of the
  incorrectly determined ROI.\label{fig:roidetection}}
\end{figure}

In order to evaluate the image processing pipeline proposed in this work,
approximately 100~image stacks with 52~or 60~frames per stack were processed.
The processing pipeline comprises a user interface which is presented to the
user at the end of the analysis and allows to review the results. The
segmentation results were verified by an expert biologist who rejected image
stacks where the first or second image processing step failed. 

The first processing step (segmentation of the nuclei) was successful in almost
99\,\% of all cases. Out of the correctly segmented image stacks, the second
processing step (identifying the site of irradiation and definining the ROI)
was accurate in 99\,\% of all cases.

It should be noted that the 100 image stacks included into the analysis also
comprise very difficult segmentation scenarios, e.g. cases where the cell moves
significantly during image acquisition, or where confounding factors such as
air bubbles or other cells (albeit without lesion) are present in the image. A
selection of difficult segmentation scenarios where our approach was still
mostly successful is shown in Figure~\ref{fig:examples}.  Each row shows sample
images that were acquired during the experiment and the avg-$t$-projection
image of the segmented and registered image stack.  Row {\it (a)} shows a very
good example where the accumulation is clearly visible and the cell hardly
moves.  Rows {\it (b)} and {\it (c)} are difficult to segment and to measure
due to cell movement and deformation of the nucleus {\it (b)} or hardly
detectable accumulation {\it (c)}.  Row {\it (d)} is another difficult example
where the intensity within the nucleus is so high that our algorithm detects a
wrong ROI.  The ROI detection for all of these image stacks is shown in
Figure~\ref{fig:roidetection}.  The top row shows the avg-$t$-projection and
the detected ROI is overlayed.  The bottom row shows the corresponding
intensity profile, the peak that the scoring algorithm chooses is clearly
marked.  Note that the ROI detection in image {\it (d)} failed, the actual
accumulation takes place to the left of the detected ROI; this is the only
case of all processed image stacks where the ROI detection failed.

The processing time per image stack is in the range of $1$--$2$ minutes per
stack on a computer running a 2.83 GHz processor. In the whole processing
pipeline, the registration and preprocessing stages are the most time consuming
steps.

\subsection{Manual vs. Automated Analysis}
\label{sec:comparison}

\begin{figure}
  \begin{center}
    \includegraphics[width=\columnwidth]{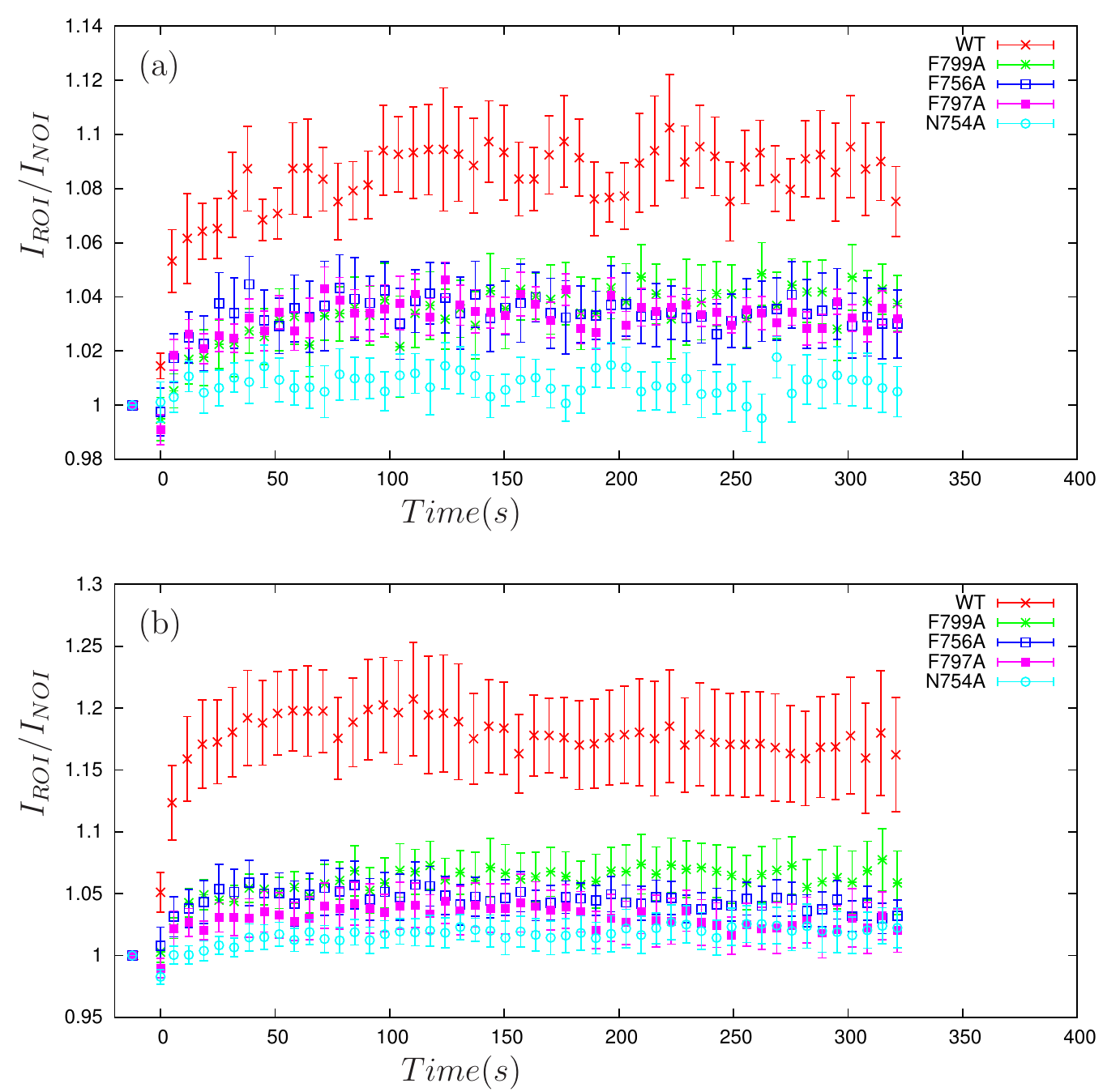}
  \end{center}
  \caption{Results of the automatic measurement {\it (a)} and the manual measurement
  {\it (b)}.  Shown are the mean values and their standard errors of the intensity
  quotient $I_{ROI}/I_{NOI}$ of several image stacks for five different XPC
  mutants in CHO cells.\label{fig:measurement}}
\end{figure}

\begin{figure}
  \begin{center}
    \includegraphics[width=\columnwidth]{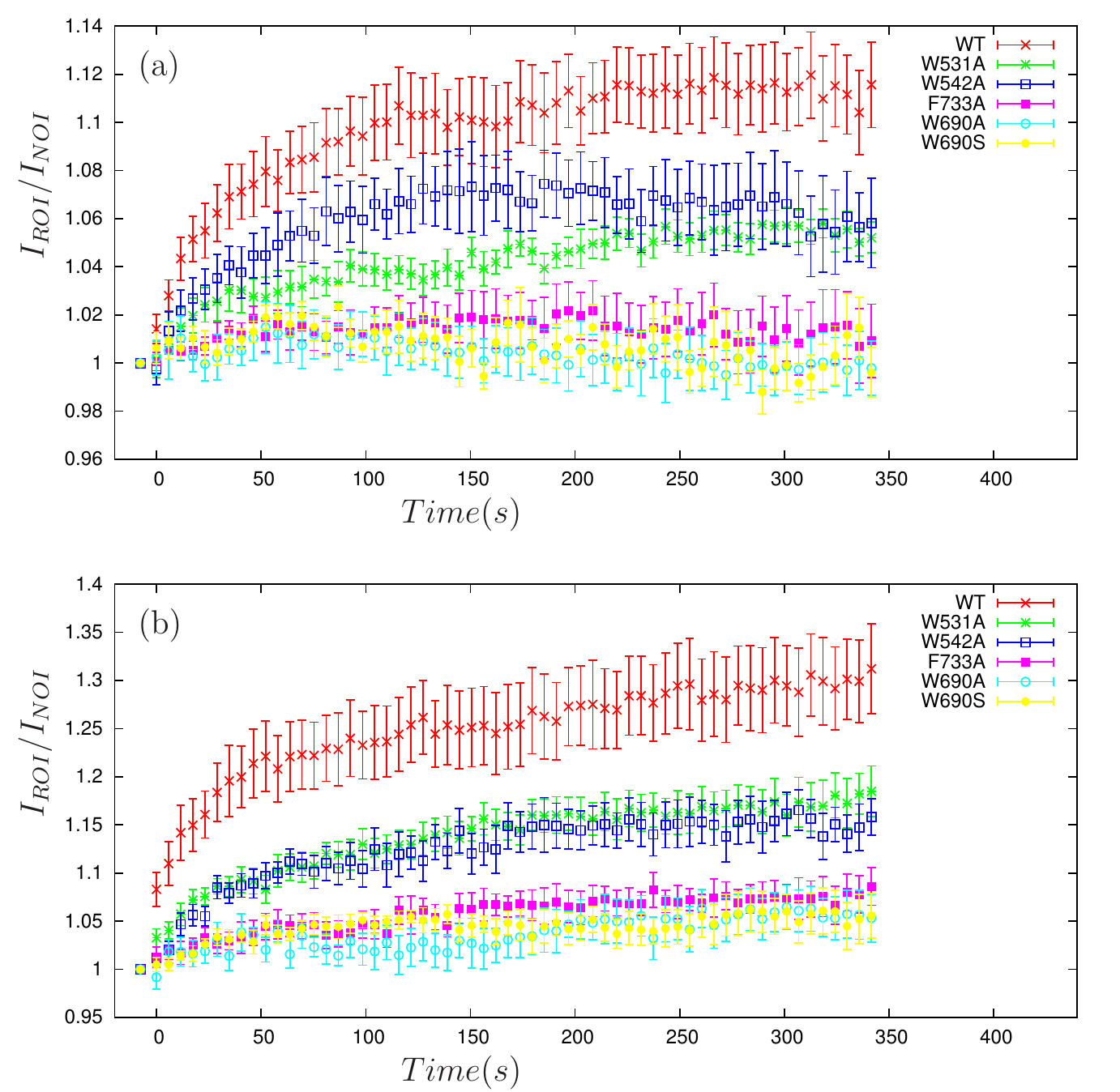}
  \end{center}
  \caption{Results of the automatic measurement {\it (a)} and the manual measurement
  {\it (b)}.  Shown are the mean values and their standard errors of the intensity
  quotient $I_{ROI}/I_{NOI}$ of several image stacks for different XPC
  mutants expressed in XP-C fibroblasts.\label{fig:measurement2}}
\end{figure}

For a more detailed assessment of the results of the automated image processing
approach, the automated measurements are compared to the manual evaluation
in~\cite{Camenisch09} and~\cite{Clement10}.  The manual evaluation is done
similar to the automated evaluation.  First, the image stack is registered,
then the ROI is determined manually and the intensity of the ROI in each image
of the stack is measured and the quotient $I_{ROI}/I_{NOI}$ is determined.  In
contrast to the automated evaluation, the ROI size is not fixed, but adjusted
to fit the most visible part of the lesion.

Figure~\ref{fig:measurement} {\it (a)} shows the result of our automatic
intensity measurements for four different XPC mutants (F756A, F797A, F799A,
N754A) and wild-type XPC expressed in CHO cells.  The quotient
$I_{ROI}/I_{NOI}$ is plotted versus the time of acquisition.  Shown are the
mean values of at least $8$ image stacks per mutant, the error bar is given by
the standard error of these mean values.  The computed results are compared to
those obtained by manual evaluation~\cite{Clement10} as shown in
Figure~\ref{fig:measurement} {\it (b)}.  Note that the graphs show the same
qualitative behaviour, but there is a quantitative difference.  The intensity
increase determined by the automated approach is much lower.  However, at the
same time there is a reduction of the standard error, hence preserving e.g. the
statistically significant difference between the data from XPC-WT and the
various XPC mutants (in this case, the statistical significance is simply
provided by non-overlapping standard error bars).  What is more, the
automatically measured intensity changes of the N754A mutant are lower than
intensity changes of the F-mutants.  Whether the improved ability of the
automated analysis to differentiate between the N and the F mutants has a
biological correlate is a question beyond the scope of this work and will be
investigated in future studies.

Figure~\ref{fig:measurement2} {\it (a)} shows the automatically measured
intensity changes for five different XPC mutants (W531A, W542A, F733A, W690A,
W690S) and wild-type XPC expressed in XP-C cells.  The quotient
$I_{ROI}/I_{NOI}$ is plotted versus the time, and again the mean values of
several image stacks per XPC mutant are shown.  The results of the manual
evaluation from~\cite{Camenisch09} are shown in Figure~\ref{fig:measurement2}
{\it (b)}; note that the graphs show the same quantitative differences as the
graphs in Figure~\ref{fig:measurement}.  The qualitative features are largely
preserved, with the exception that the automatically measured values of W531A
and W542A are separated in the time interval starting at $50s$ and ending at
$220s$.  This separation cannot be seen in the manual evaluation and it is not
entirely clear if there are biological reasons for this separation or if it is
due to the image quality of the W542A mutants, which is also expressed in the
high standard error in the automated analysis result.

Summarizing, the results of the automated measurements preserve the features
that were discovered in the manual evaluation, and on top of that have the
advantage of reproducibility and unbiasedness, and, last but not least, time
saving.  Moreover, the measurement seems to be more accurate, the N754A mutant
appears clearly separated from the F-mutants in the experiments involving the
CHO cells, which is not the case for the manual evaluation.

\section{Discussion}

The results of our automated image processing approach are very satisfactory.
We achieve almost 99\,\% correctly segmented nuclei and 99\,\% correctly
determined ROIs for our data set of approximately 100 image stacks included in
the analysis.  Moreover, the comparison with a manual evaluation of the data
set shows that the automated measurement not only supports the qualitative
statements that can be drawn from the manual evaluation, but also has a lower
standard error.  In the presented case, the automated evaluation shows that the
N754A mutants behave significantly different from the other evaluated mutants.
Whether the improved ability of the automated analysis to differentiate between
the N and the F mutants has a biological correlate is a question beyond the
scope of this work and will be investigated in future studies.  

The quantitative differences of the measured intensity values in the manual and
in the automated evaluation are likely due to the choice of the ROI size.  In
the manual evaluation, the ROI has been chosen smaller for the XPC mutants
where the XPC-GFP accumulation is clearly visible, and larger for the XPC
mutants where hardly any accumulation is visually detectable.  This choice
influenced the measured intensity changes, in particular the values for mutants
with high accumulation are much higher than in the automated analysis, where
the ROI has a constant size throughout the analysis.  Although this
automated approach performs very well for our experimental data sets, there is
room for improvements regarding the registration algorithm --- a nonlinear
registration algorithm~\cite{Kim09} can also compensate deformation of the
irradiated area --- and computation time.

\section{Conclusion}

We present an automated system for measuring the performance of XPC in the DNA
repair process based on intensity changes in microscopy images.  The image
processing pipeline comprises several steps that are based on standard image
processing algorithms in combination with a customized segmentation
algorithm and a specific scoring method to detect the correct ROI.

Laser microirradiation in combination with fluorescence microscopy has become a
popular method for studying the dynamics of DNA repair in live cells.
Computational tools that facilitate the extraction of quantitative data from
such experiments are  therefore of great interest to the biology community.
Further work will be directed at recognizing more complex irradiation and damage
patterns.

\section{Acknowledgments}

The  Interdisciplinary Center for Interactive Data Analysis, Modelling and
Visual Exploration (INCIDE) is funded via a grant of the German Excellence
Initiative by the German Research Foundation (DFG) and the German Council of
Science and Humanities awarded to the University of Konstanz.  The Center of
Applied Photonics (CAP) is supported by the Ministry of Science, Research and
the Arts Baden-W\"urttemberg.  We thank D.~Tr\"autlein and H.~Naegeli for
providing image data.  We gratefully acknowledge A. Leitenstorfer, M. Horn and
O. Deussen for fruitful discussions and support.






\bibliographystyle{ieeetr}
\bibliography{literature}

\end{document}